\documentclass[letterpaper, 10 pt, conference]{ieeeconf}  % Comment this line out if you need a4paper

\IEEEoverridecommandlockouts                              % This command is only needed if 
\usepackage{graphicx} % for PDF, bitmapped graphics files
\usepackage{amsmath} % assumes amsmath package installed
\usepackage{amssymb}  % assumes amsmath package installed
\usepackage{booktabs} % For professional looking tables
\usepackage{caption}  % For table captions
\usepackage{siunitx}  % For aligning numbers by decimal point
\usepackage{cite}

\makeatletter
\let\NAT@parse\undefined
\makeatother
\usepackage{hyperref}

\hypersetup{
    colorlinks=true,   
    linkcolor=blue,    
    urlcolor=magenta,  
    citecolor=blue   
}

\title{\LARGE \bf
MVAdapt: Zero-Shot Multi-Vehicle Adaptation 

for End-to-End Autonomous Driving
}

\author{Haesung Oh$^{1}$ and Jaeheung Park$^{2}$% <-this % stops a space
\thanks{$^{1}$Haesung Oh is with the Interdisciplinary Program in AI, Seoul National University, South Korea
        {\tt\small haesungoh@snu.ac.kr}}%
\thanks{$^{2}$Jaeheung Park with the Faculty of the Interdisciplinary Program in AI, Seoul National University, South Korea
        {\tt\small park73@snu.ac.kr}}%
}

\begin{document}

\maketitle
\thispagestyle{empty}
\pagestyle{empty}

%%%%%%%%%%%%%%%%%%%%%%%%%%%%%%%%%%%%%%%%%%%%%%%%%%%%%%%%%%%%%%%%%%%%%%%%%%%%%%%%
\begin{abstract}
End-to-End (E2E) autonomous driving models are usually trained and evaluated with a fixed ego-vehicle, even though their driving policy is implicitly tied to vehicle dynamics. When such a model is deployed on a vehicle with different size, mass, or drivetrain characteristics, its performance can degrade substantially; we refer to this problem as the \emph{vehicle-domain gap}. To address it, we propose \emph{MVAdapt}, a physics-conditioned adaptation framework for multi-vehicle E2E driving. MVAdapt combines a frozen TransFuser++ scene encoder with a lightweight physics encoder and a cross-attention module that conditions scene features on vehicle properties before waypoint decoding. In the CARLA Leaderboard 1.0 benchmark, MVAdapt improves over naive transfer and multi-embodiment adaptation baselines on both in-distribution and unseen vehicles. We further show two complementary behaviors: strong zero-shot transfer on many unseen vehicles, and data-efficient few-shot calibration for severe physical outliers. These results suggest that explicitly conditioning E2E driving policies on vehicle physics is an effective step toward more transferable autonomous driving models. All codes are available at \url{https://github.com/hae-sung-oh/MVAdapt}

\end{abstract}

%%%%%%%%%%%%%%%%%%%%%%%%%%%%%%%%%%%%%%%%%%%%%%%%%%%%%%%%%%%%%%%%%%%%%%%%%%%%%%%%
\section{Introduction}
Conventional autonomous driving divides the pipeline into several modules, such as perception, localization, planning, and control. In contrast, End-to-end (E2E) autonomous driving aims to learn a single AI model that directly maps raw sensor data inputs to vehicle control commands. The approach has an advantage over the conventional way for preventing error propagation through modules by optimizing the entire system jointly. In this regard, it solves the primary hurdle of traditional modular pipelines \cite{c1, c2, c3, c4, c5}. However, generalization capability to untrained domains remains a significant barrier. Recent works have focused on environment domain adaptations, such as sim-to-real adaptations \cite{c6, c7, c8, c9}, climate adaptations \cite{c10, c11}, and region adaptations \cite{c12, c13, c14, c15}. In this paper, we introduce a novel domain adaptation paradigm, which we term \emph{vehicle-domain adaptation}, as suggested by \cite{c16} for its necessity. This research perspective has been overlooked because current E2E models treat the mapping between vehicle dynamics and vehicle maneuver as part of the hidden, implicit feature to be approximated, rather than as an explicit factor to be learned.

\begin{figure}[t!!]
  \centering
    \includegraphics[width=\columnwidth]{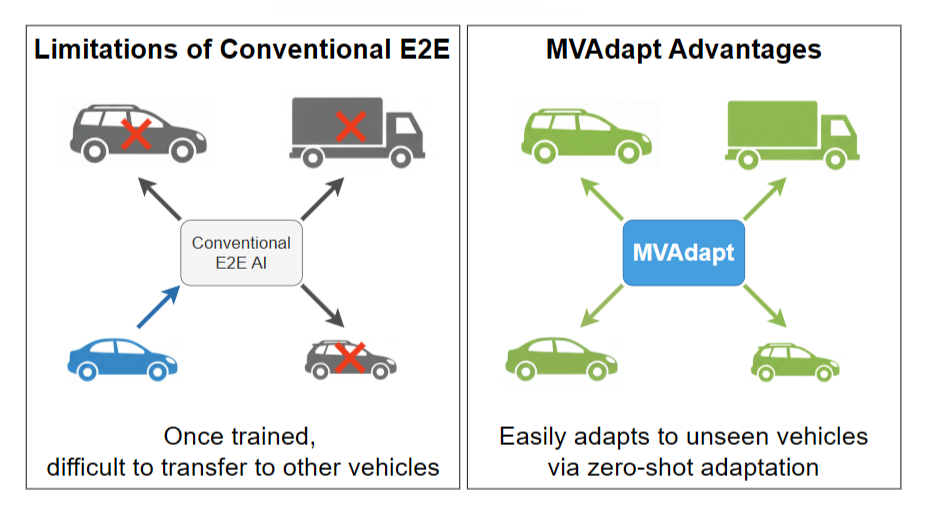}
  \caption{\emph{Advantages of MVAdapt}: Conventional E2E models lack transferability across vehicles (left), while MVAdapt enables zero-shot adaptation to unseen vehicle types (right).}
  \label{intro1}
\end{figure}

\begin{figure}[t!!]
  \centering
    \includegraphics[width=\columnwidth]{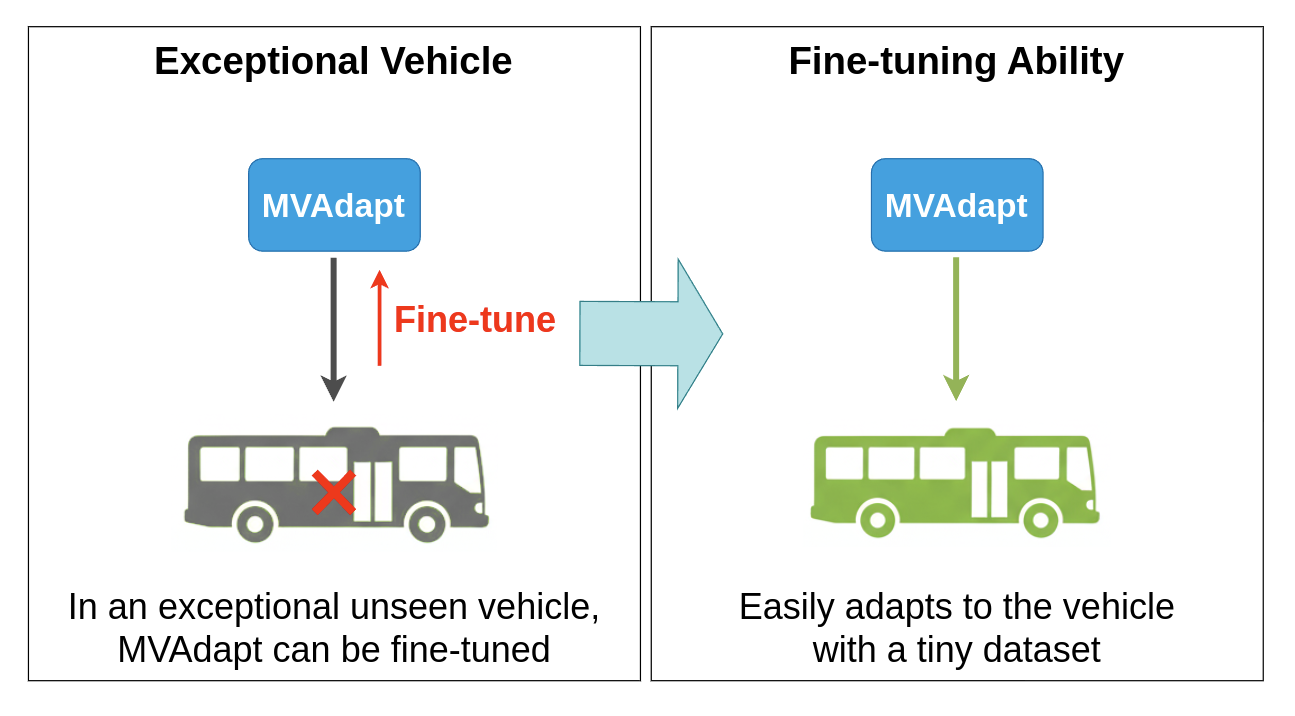}
  \caption{\emph{Few-shot Adaptation}: Even if MVAdapt is not able to adapt to an exceptional unseen vehicle in a zero-shot manner (left), it shows fine-tuning ability with a minimal dataset (right).}
  \label{intro2}
\end{figure}

E2E models learn an implicit mapping from perception to control, which inherently embeds vehicle-specific dynamics. For instance, an E2E model trained on a lightweight sedan encodes a sedan-specific driving maneuver by learning the expert actions for the vehicle. When it is naively transferred to a big and heavy SUV, the same driving outputs with the same sensor inputs cause critical dangers. First, an SUV and a sedan will exhibit distinct dynamic responses even when subjected to identical control inputs. Second, trajectories considered safe for the sedan may be infeasible or hazardous for the SUV. Lastly, the model may generate a command that is kinematically infeasible for the SUV to execute. 

Especially, the process of data collection and model retraining for each target vehicle is prohibitively resource-intensive, hindering the widespread deployment of E2E models. Therefore, effective vehicle-domain adaptation must be achieved by internalizing the physical attributes of vehicles and the requisite adjustments in their driving policy. In our evaluation, this gap is large enough to be practically consequential: a TransFuser++ waypoint model trained for the default vehicle drops to average Driving Scores of 19.31 on the 27 training-distribution vehicles and 28.77 on 31 unseen vehicles when naively transferred, far below its vehicle-specific performance on the source car. These results motivate vehicle-domain adaptation as a concrete transfer problem rather than only a conceptual one.

% \newpage
To address this vehicle domain gap, we propose \textbf{MVAdapt}, a physics-conditioned adaptation framework for across-vehicle autonomous driving. 

Our contributions are as follows:

\begin{itemize}
    \item We introduce and quantify the \emph{vehicle-domain gap} in end-to-end autonomous driving, showing that naive transfer across vehicle embodiments causes a severe performance drop even when the visual driving task is unchanged.
    \item We propose \textbf{MVAdapt}, a cross-attention-based adaptation module that combines frozen scene features with normalized vehicle physics. The method is designed for strong zero-shot transfer on many unseen vehicles and data-efficient few-shot calibration on severe outliers.
    \item Experiments on the CARLA Leaderboard 1.0 benchmark show that MVAdapt consistently improves over naive transfer and adapted multi-embodiment robotics baselines, while the ablation study confirms that both the physics encoder and the cross-attention fusion contribute to the gain.
\end{itemize}

\begin{figure*}[ht]
  \centering
    \includegraphics[width=\textwidth]{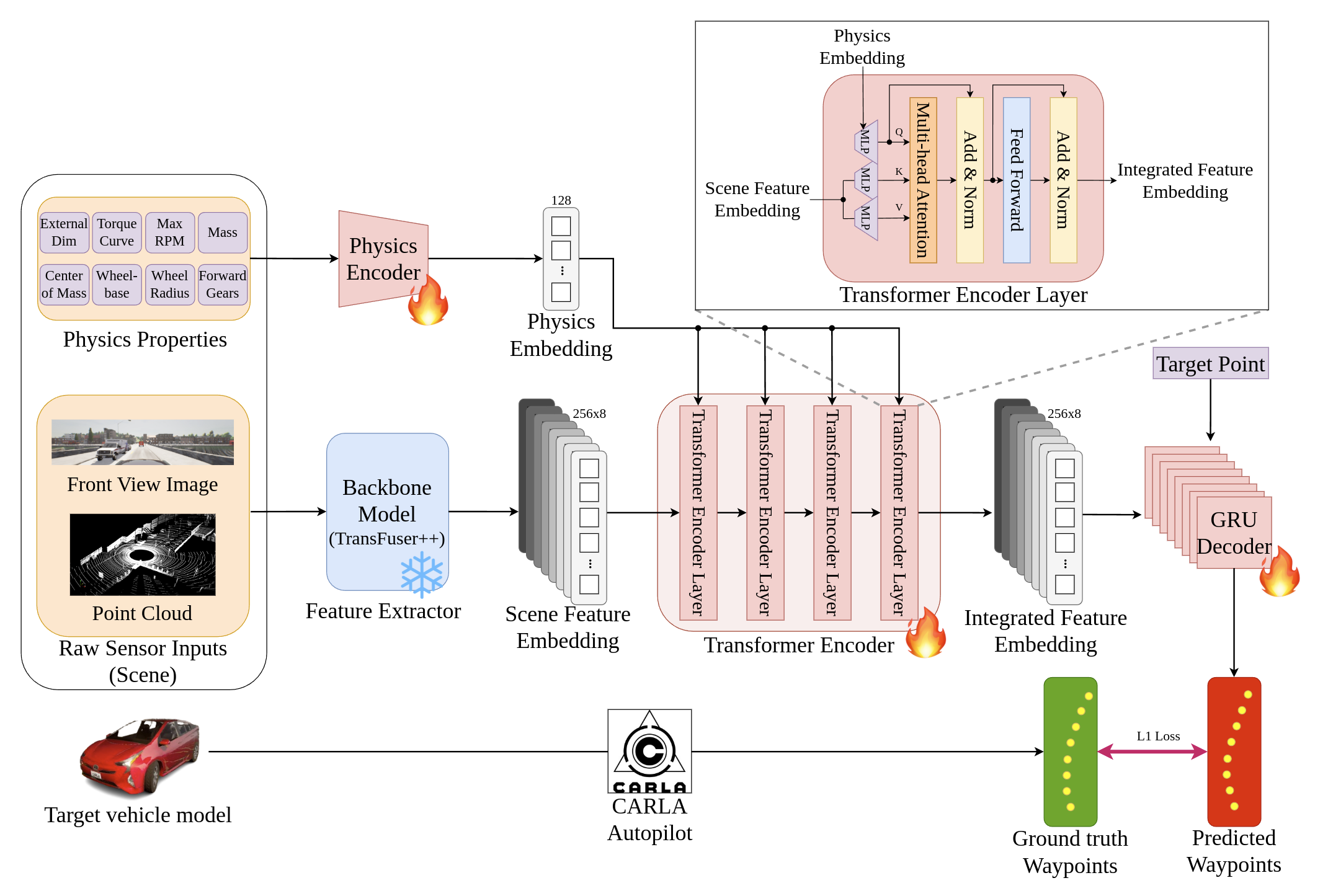} 
  \caption{\emph{Overall architecture of MVAdapt}: Raw sensor inputs (camera image and LiDAR point cloud) are processed by a frozen TransFuser++ backbone to extract scene features, while vehicle-specific physical properties are encoded into a physics embedding. A multi-head transformer encoder fuses the physics embedding with scene features, producing an integrated feature embedding that conditions perception on the ego-vehicle’s dynamics. Finally, a GRU-based decoder generates future waypoints toward the target point, trained with an \(L_1\) loss against expert trajectories from CARLA autopilot.}
  \label{fig:mvadapt_arch}
\end{figure*}

The remainder of the paper is structured as follows: Section \ref{sec:works} outlines related previous works. Afterward, Section \ref{sec:method} derives the framework MVAdapt. The validation experiments are presented in Section \ref{sec:experiments}. After discussing the results in Section \ref{sec:results}, Section \ref{sec:conclusion} summarizes the findings and suggests directions for future work.

\section{Related Works} \label{sec:works}

\subsection{End-to-End Autonomous Driving}

The field has evolved from early CNN-based models such as PilotNet \cite{c17, c18} to more sophisticated architectures. More recently, transformer-based models such as TransFuser \cite{c19, c20} and InterFuser \cite{c21} have set new performance benchmarks by effectively fusing multi-sensor data. Despite these advances, most research implicitly assumes a fixed vehicle embodiment, and the problem of generalization across different vehicle dynamics remains largely unaddressed.

\subsection{Domain Adaptation in End-to-End Autonomous Driving}

A significant body of research has focused on bridging other domain gaps, such as the sim-to-real gap \cite{c6, c7, c8, c9} or adapting to varied environmental conditions like climate \cite{c10, c11} and region \cite{c12, c13}. To support such domain shift problems, a parallel line of work has conducted dedicated datasets \cite{c22, c23}. These methods typically focus on achieving visual feature invariance and do not provide mechanisms to adapt the control policy to changes in the agent’s underlying physical properties.

\subsection{Physics-Informed Learning in End-to-End Autonomous Driving}

Another line of research seeks to integrate physical knowledge into learning-based models. This field includes hybrid approaches that couple deep learning perception modules with classical controllers like MPC \cite{c24, c25, c26}. More autonomous driving works based on Physics-Informed Neural Networks (PINNs) exist, such as Deep Dynamics \cite{c27} and FusionAssurance \cite{c28}, which learn to generate physics-constrained maneuvers. In contrast, MVAdapt’s objective is to condition the policy on known a priori physical parameters, enabling immediate, zero-shot transfer.

\subsection{Multi-Embodiment Robotics Control}

The problem of creating a single policy for multiple robot morphologies is well-studied in legged robotics. Various strategies have been proposed to tackle this challenge, with recent efforts often leveraging architectural innovations to encode morphological information \cite{c29, c30, c31}.

\emph{Unified Robot Morphology Architecture (URMA)} \cite{c32} uses an attention mechanism to create a fixed-size latent representation from varied joint observations and descriptions, enabling control of different robot types.

\emph{Body Transformer (BoT)} \cite{c33}  represents the robot’s body as a graph and uses masked attention to leverage the physical structure as an inductive bias. These approaches provide the closest conceptual baseline family for our setting because they explicitly condition control policies on embodiment information. We therefore adapt and implement these two as robust comparison methods, translating their discrete, joint-based representations to the continuous parameter space of vehicle physics.

\subsection{Vehicle-Specific Dynamics Modeling}

Recent works have begun to address vehicle dynamics more directly, though with different objectives than our own. 

\emph{AnyCar to Anywhere} \cite{c34} proposes a universal dynamics model for agile control, demonstrating impressive generalization for trajectory tracking tasks after fine-tuning on real-world data. However, its focus remains on dynamic prediction for agile mobility rather than learning a complete, reactive policy for complex urban driving scenarios. 

Similarly, \emph{One Model to Drift Them All} \cite{c35} successfully adapts a single model for extreme, at-the-limit maneuvers on two distinct real-world vehicles. Yet, its scope is intentionally limited to the specialized domain of drifting, not general on-road navigation, and it relies on online adaptation rather than achieving accurate zero-shot transfer to unseen vehicles. 

Closer to our problem formulation, \emph{Vehicle Type Specific Waypoint Generation} \cite{c36} presents a method to make a general behavioral model produce more physically plausible waypoints for a specific vehicle type. While it considers vehicle properties, it addresses waypoint generation in specific turning scenarios only, rather than creating an integrated, end-to-end urban driving policy. These approaches highlight the importance of vehicle dynamics, but also underscore the novelty of MVAdapt’s goal: achieving robust zero-shot adaptation for a complete end-to-end (E2E) driving policy in general traffic scenarios.

\section{Methodology} \label{sec:method}

The MVAdapt architecture (Fig.~\ref{fig:mvadapt_arch}) is designed to integrate physical properties into an existing E2E driving model. It consists of a frozen pre-trained feature extractor and a physics-attention adaptation module. 

\subsection{Data Representation and Preprocessing}
MVAdapt takes two input streams: multi-modal scene observations and vehicle physics. RGB images are processed with the same normalization pipeline as the TransFuser++ backbone, while LiDAR point clouds are converted into the bird's-eye-view representation used by the waypoint model. For the physics branch, scalar attributes with very different ranges (e.g., mass, wheel radius, and center of mass) are normalized to a common range before fusion. Variable-length attributes such as torque-curve points and forward-gear parameters are zero-padded to fixed size so that the encoder receives a consistent input dimension across vehicle types. This preprocessing makes the adaptation module less sensitive to raw scale differences between physical parameters and keeps the fusion stage numerically stable across diverse vehicle types.

\subsection{Backbone Feature Extractor} 
To ensure robust multi-modal perception, we employ \emph{TransFuser++ WP} as the backbone feature extractor \cite{c20}. More specifically, we use the fused waypoint-prediction features after multi-modal image--LiDAR fusion and before the original TransFuser++ GRU decoder. The resulting scene representation is treated as a sequence of $N_{target}=8$ high-level scene tokens, each aligned with one future prediction step. These backbone weights are pre-trained on CARLA using the \texttt{Lincoln MKZ 2017} vehicle model and are kept frozen throughout MVAdapt training. Freezing the scene encoder isolates vehicle adaptation to the lightweight conditioning module and makes it clear that the performance gain does not come from re-learning perception from scratch.

\subsection{Physical Properties and Physics Encoder}

We collect vehicle properties from the CARLA API (Table~\ref{tab:physical_properties}), flatten them into a single vector, and preprocess them as described above. The normalized physics vector is then passed through a small MLP-based encoder to produce a compact physics embedding. In practice, this embedding summarizes both expected vehicle behavior (e.g., how aggressively the vehicle can accelerate or turn) and operational constraints (e.g., feasible turning radius or inertial burden). Compared with directly concatenating raw physical values to the decoder input, this encoder provides a cleaner latent representation for downstream fusion.

\subsection{Multi-Head Transformer Encoder}

We fuse scene representations with vehicle physics through a multi-head attention mechanism. Let $z_{scene}\in\mathbb{R}^{N_{target}\times d}$ denote the sequence of scene tokens and $z_{phys}\in\mathbb{R}^{d}$ denote the encoded vehicle physics. We project the physics embedding to the query and the scene tokens to keys and values, so that the vehicle embedding attends to the most relevant scene elements for the current embodiment. Concretely, larger vehicles can emphasize clearance-related cues, while lighter or shorter-wheelbase vehicles can place more weight on tighter maneuver opportunities. In the implementation used for the thesis appendix, this fusion block uses 4 transformer layers with 8 attention heads and a feed-forward dimension of 512. This attention-driven fusion produces a physics-informed scene representation that is passed to the waypoint decoder.

\begin{table*}[h!]
\centering
\vspace*{10pt}
\caption{Physical Properties for MVAdapt}
\label{tab:physical_properties}
\begin{tabular}{lll}
\toprule
\textbf{Physical Property} & \textbf{Dimension} & \textbf{Description} \\
\midrule
External Dimension & $3$ & The overall size of the vehicle in meters (length, width, height). \\
Torque Curve & $2 \times N_{\text{torque}}$ & Data points representing engine torque (Nm) at various RPMs. \\
Max RPM & $1$ & The maximum rotational speed the engine can achieve. \\
Mass & $1$ & The total mass of the vehicle. \\
Center of Mass & $3$ & The (x, y, z) coordinates of the vehicle’s center of gravity. \\
Wheelbase & $1$ & The longitudinal distance between the front and rear axles. \\
Wheel Radius & $1$ & The radius of the vehicle’s wheels. \\
Forward Gears & $3 \times N_{\text{gears}}$ & Defines each gear by its ratio, up-shift RPM ratio, and down-shift RPM ratio. \\
\bottomrule
\end{tabular}
\vspace{-10pt}
\end{table*}

\subsection{Output Head}

The output from the adaptation module is conceptualized as an integrated scene feature embedding. This feature is subsequently fed into the final output head, which consists of a Gated Recurrent Unit (GRU). The GRU also takes the coordinates of the target point, which is sequentially given by the CARLA Leaderboard benchmark, as an additional input. To provide target points to the model in the real world, several path planning algorithms \cite{c39, c40, c41} might be needed. The GRU iterates through the stack of \(N_{target} = 8\) scene feature embeddings to sequentially generate \(N_{target}\) time-variant waypoint coordinates. This sequence of waypoints constitutes the future trajectory of the ego vehicle. Finally, this predicted trajectory is passed to a fixed low-level PID controller, which generates the control inputs (i.e., throttle, steering, and brake).

\section{Experiments} \label{sec:experiments}
\subsection{Experimental Setup}
\subsubsection{Simulation Environment}
We use the CARLA simulator \cite{c37} version 0.9.12, a standard tool for autonomous driving research. It provides an open-source platform built to support the development, training, and validation of autonomous systems. To standardize evaluation, the CARLA Autonomous Driving Leaderboard \cite{c38} is used. The benchmark challenges agents to navigate predefined routes across diverse environments while handling a variety of complex traffic scenarios.

\subsubsection{Data Generation}
With the default vehicle catalogue of the CARLA simulation, we can collect the ground truth driving datasets for 27 distinct vehicle models. For each vehicle archetype, a rule-based autopilot of the simulation generates driving datasets, which include sensor data, corresponding trajectory, and the physical properties of the vehicle. To ensure the quality of the supervising data, we selected the perfect driving scenarios without any accidents. Additionally, a huge truck and a tiny car are excluded to validate the adaptation ability to extreme unseen vehicles. 

\subsubsection{Benchmark and Metrics}
We evaluate the driving performance of MVAdapt with all vehicles on the \texttt{longest6} benchmark, which consists of 6 routes in 6 different towns. Performance is measured using the official CARLA Leaderboard metrics:

\begin{itemize}
    \item \textbf{Route Completion (RC)}: The percentage of the route that the agent completed before the termination of the scenario due to a high-risk accident (e.g., collision, rollover, off-road, stuck, etc.).
    \begin{equation}
        RC = \frac{l_{complete}}{l_{total}}
    \end{equation}
    \item \textbf{Infraction Score (IS)}: A penalty factor that decreases from 1.0 for every traffic violation (e.g., collision, ignoring a stop sign, traffic light violations, etc.)
    \begin{equation}
        IP = \prod_{k} p_{k}
    \end{equation}
    \begin{equation*}
    p_k \sim (0,\ 1)\ \text{on\ every\ violation}
    \end{equation*}
    \begin{equation*}
        p_k = \begin{cases}
        0.5 & \text{Collisions\ with\ pedestrians} \\
        0.6 &  \text{Collisions\ with\ other\ vehicles} \\
        0.65 & \text{Collisions\ with\ static\ elements} \\
        0.7 & \text{Running\ a\ red\ light} \\
        0.8 & \text{Running\ a\ stop\ sign}
    \end{cases}
    \end{equation*}
    \item \textbf{Driving Score (DS)}: The primary metric, calculated as the product of Route Completion and Infraction Score.
    \begin{equation}
        DS = RC_{i}IP_{i}
    \end{equation}
\end{itemize}

\subsection{Training}
The model is trained on a dataset comprising 27 vehicles, which contains approximately 1 million frames in total. This training procedure maps the multi-modal sensor data and the future waypoints. \(L1\) loss is used between the predicted waypoints and the rule-based expert’s waypoints. 

\subsection{Zero-Shot Adaptation}
To analyze the zero-shot adaptation ability of the proposed method, we deployed the trained AI model to several unseen vehicle models. They contain not only the excluded vehicles from the CARLA catalogue, but also vehicles with stochastically distributed physics. The physical properties of these \emph{‘sampled vehicles’} are modified with the CARLA API, so that the vehicles act based on the sampled physics. To ensure realistic maneuvers, the sampling range for each physical parameter was carefully determined. This experimental setup has enabled a more diverse and flexible range of zero-shot adaptation experiments.

\subsection{Few-Shot Adaptation with Fine-tuning}
Additionally, we conduct a few-shot adaptation experiment: we fine-tune MVAdapt on a tiny dataset ($\sim$ 37K frames, $\approx$ 3\% of the full training) of the \texttt{Carla Cola Truck} driving, then evaluate its performance on that truck. This experiment simulates a scenario in which a new vehicle becomes available, and we update the model using minimal data. We examine how quickly performance improves on the vehicle.

\begin{table*}[ht]
\vspace*{10pt}
\centering
\caption{In-distribution Vehicles (Average over 27 vehicles)}
\label{tab:in-distribution-results}
\begin{tabular}{lccc}
\toprule
\textbf{Model} & \textbf{Avg. Driving Score (DS) $\uparrow$} & \textbf{Avg. Route Completion (RC) $\uparrow$} & \textbf{Avg. Infraction Score (IS) $\uparrow$} \\
\midrule
TransFuser++ (Default Vehicle) & 80.27 & 96.33 & 0.83 \\
\midrule
TransFuser++ (Naive Transfer)  & 19.31 & 46.85 & 0.45 \\
URMA                           & 33.25 & 46.63 & 0.71 \\
BodyTransformer                & 36.92 & 58.86 & 0.56 \\
\textbf{MVAdapt (Ours)}        & \textbf{78.21} & \textbf{96.62} & \textbf{0.80} \\
\bottomrule
\end{tabular}
\end{table*}

\begin{table*}[ht]
\centering
\caption{Out-of-distribution Vehicles (Average over 31 vehicles)}
\label{tab:out-of-distribution-results}
\begin{tabular}{lccc}
\toprule
\textbf{Model} & \textbf{Avg. Driving Score (DS) $\uparrow$} & \textbf{Avg. Route Completion (RC) $\uparrow$} & \textbf{Avg. Infraction Score (IS) $\uparrow$} \\
\midrule
TransFuser++ (Naive Transfer)  & 28.77 & 54.23 & 0.51 \\
URMA                           & 24.39 & 46.83 & 0.48 \\
BodyTransformer                & 23.21 & 51.06 & 0.43 \\
\textbf{MVAdapt (Ours)}        & \textbf{63.02} & \textbf{96.77} & \textbf{0.65} \\
\bottomrule
\end{tabular}
\end{table*}

\begin{table*}[ht]
\centering
\caption{Fine-Tuning on unseen \texttt{Carla Cola Truck}}
\label{tab:finetuning-overall-results}
\begin{tabular}{lccc}
\toprule
\textbf{Model} & \textbf{Avg. Driving Score (DS) $\uparrow$} & \textbf{Avg. Route Completion (RC) $\uparrow$} & \textbf{Avg. Infraction Score (IS) $\uparrow$} \\
\midrule
\multicolumn{4}{l}{\textbf{Performance on \texttt{Carla Cola Truck}}} \\
MVAdapt (Unseen, before FT) & 30.37 & 100.0 & 0.30 \\
\textbf{MVAdapt (Fine-tuned)} & \textbf{61.90} & \textbf{100.0} & \textbf{0.62} \\
% \midrule
% \multicolumn{4}{l}{\textbf{Overall performance on all vehicles}} \\
% MVAdapt (Vanilla, before FT) & 78.21 & 96.62 & 0.80 \\
% MVAdapt (Fine-tuned) & 71.98 & 92.59 & 0.77 \\
\bottomrule
\end{tabular}
\vspace{-10pt}
\end{table*}

\subsection{Baselines}
We compare MVAdapt against three baselines:
\begin{itemize}
    \item \emph{Naive Transfer}: This is the case where we use the backbone model and its low-level decoder only (TransFuser++ waypoint model). It is trained on the source vehicle (\texttt{Lincoln MKZ 2017}) and deployed to the target vehicles without any adaptation. It represents the standard practice of training an end-to-end model per vehicle and reveals the drop in performance when the embodiment changes. This baseline helps quantify the performance gap due to vehicle mismatch.
    \item \emph{URMA-style adapter}: To compare with prior adaptation methods, we implement a variant of \emph{One Policy to Run Them All (URMA)} for the driving task. In URMA, a policy gets a description of the agent’s body (e.g., joint parameters) and uses attention to condition the action generation. We adapt this idea by providing a learned encoding of the vehicle’s properties to the policy and integrating it via a soft attention mechanism, similar to URMA’s approach. Therefore, the URMA baseline has access to the same vehicle parameters as MVAdapt, but its architectural integration has simpler attention than our full transformer fusion. It also samples stochastic actions during training, as in the original URMA, for robustness.
    \item \emph{BodyTransformer-style adapter}: We also adapt the \emph{BodyTransformer} baseline to our setting. \emph{BodyTransformer} introduced a way to integrate structured information about an agent’s body as a graph of joints into a transformer policy. This baseline is also equipped with the same vehicle properties as our method, but utilizes a different network design. The baseline treats the physical properties as a \emph{body part list} and applies a transformer encoding similar to \emph{BodyTransformer} to mix those. It provides another point of comparison for utilizing embodiment information.
\end{itemize}
MVAdapt, URMA-style, and BodyTransformer-style models are trained with the same dataset to ensure an accurate comparison.

\section{Results} \label{sec:results}
\subsection{Quantitative Performance}
\subsubsection{Performance on In-Distribution Vehicles}

MVAdapt achieves sufficient driving performance on the vehicles for which it was trained (in-distribution) using a single unified model. TABLE~\ref{tab:in-distribution-results} summarizes the results. MVAdapt achieves an average Driving Score of 78.21, outperforming the naive TransFuser++ transfer (DS 19.31) and significantly surpassing the URMA (DS 33.25) and BodyTransformer (DS 36.92) baselines. MVAdapt’s Driving Score is comparable to TransFuser++, which is specialized and trained on a specific vehicle, achieving a score of over 80 on its own vehicle. 

More impressively, MVAdapt achieves an average Route Completion of 96.92\%, meaning it can navigate the entire route without termination, most of the time. In contrast, the baselines show 47-59\% RC score on average, as they often fail to complete due to crucial accidents such as collisions or getting stuck. This route completion improvement is critical, as the proposed method can avoid high-risk mistakes. The IS of MVAdapt is 0.80, which is also the highest. It reflects fewer traffic rule violations. In summary, under in-distribution vehicles used in training, MVAdapt maintains robust, safe driving across vehicles, whereas the naive method or other adaptation approaches struggle with some vehicles.

\subsubsection{Performance on Out-of-Distribution (Unseen) Vehicles (Zero-Shot Adaptation)}
On completely unseen vehicle types, MVAdapt delivers substantially stronger zero-shot transfer than the comparison methods. As shown in Table~\ref{tab:out-of-distribution-results}, MVAdapt reaches an average DS of 63.02 on novel vehicles, compared with 28.77 for naive transfer and 24.39/23.21 for the URMA and BodyTransformer baselines. The RC remains high at 96.77\%, indicating that the model usually completes the route even under embodiment shift, while the improved IS shows fewer infractions than the alternatives. Rather than claiming universal zero-shot success for every possible vehicle, we interpret these results more narrowly: conditioning on vehicle physics materially improves transfer to a broad set of unseen vehicles in our CARLA evaluation.

Notably, the zero-shot result should be distinguished from the severe-outlier setting discussed next. On average unseen vehicles, MVAdapt maintains DS above 60 with RC above 97\%; for a much more extreme outlier such as the \texttt{Carla Cola Truck}, zero-shot transfer is weaker and few-shot calibration becomes important.

\begin{figure}
\centering
\includegraphics[width=\columnwidth]{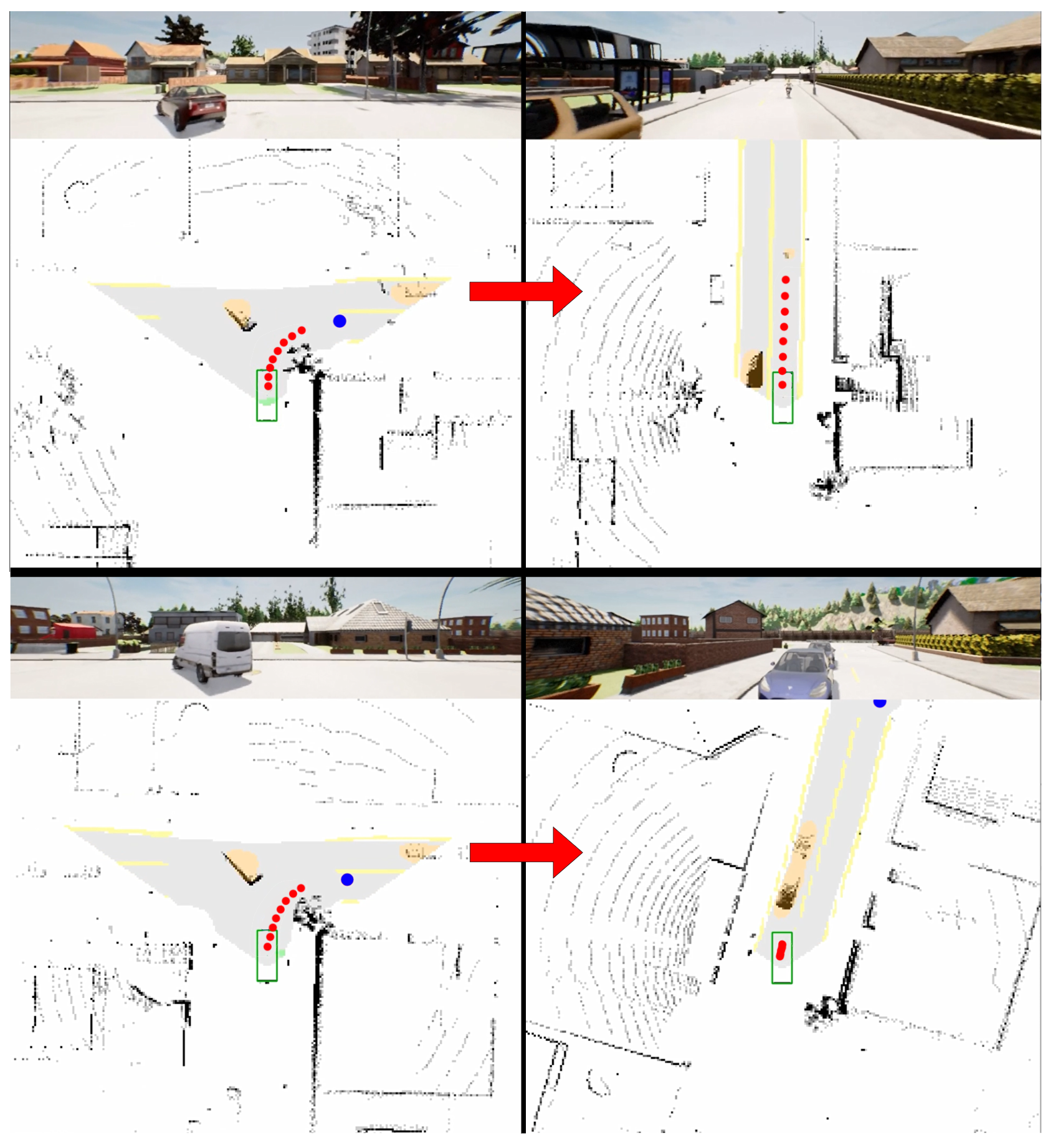}
\caption{A \texttt{Tesla Cybertruck} making a right turn. \textbf{Top (Ours)}: MVAdapt successfully navigates the turn by accounting for the vehicle’s large size. \textbf{Bottom (Baseline)}: The baseline misjudges the turning radius and gets blocked by another car. The red dots represent the model’s output trajectory, and the blue dot indicates the target point.}
\label{fig:cybertruck}
\end{figure}

\begin{figure}
\centering
\includegraphics[width=\columnwidth]{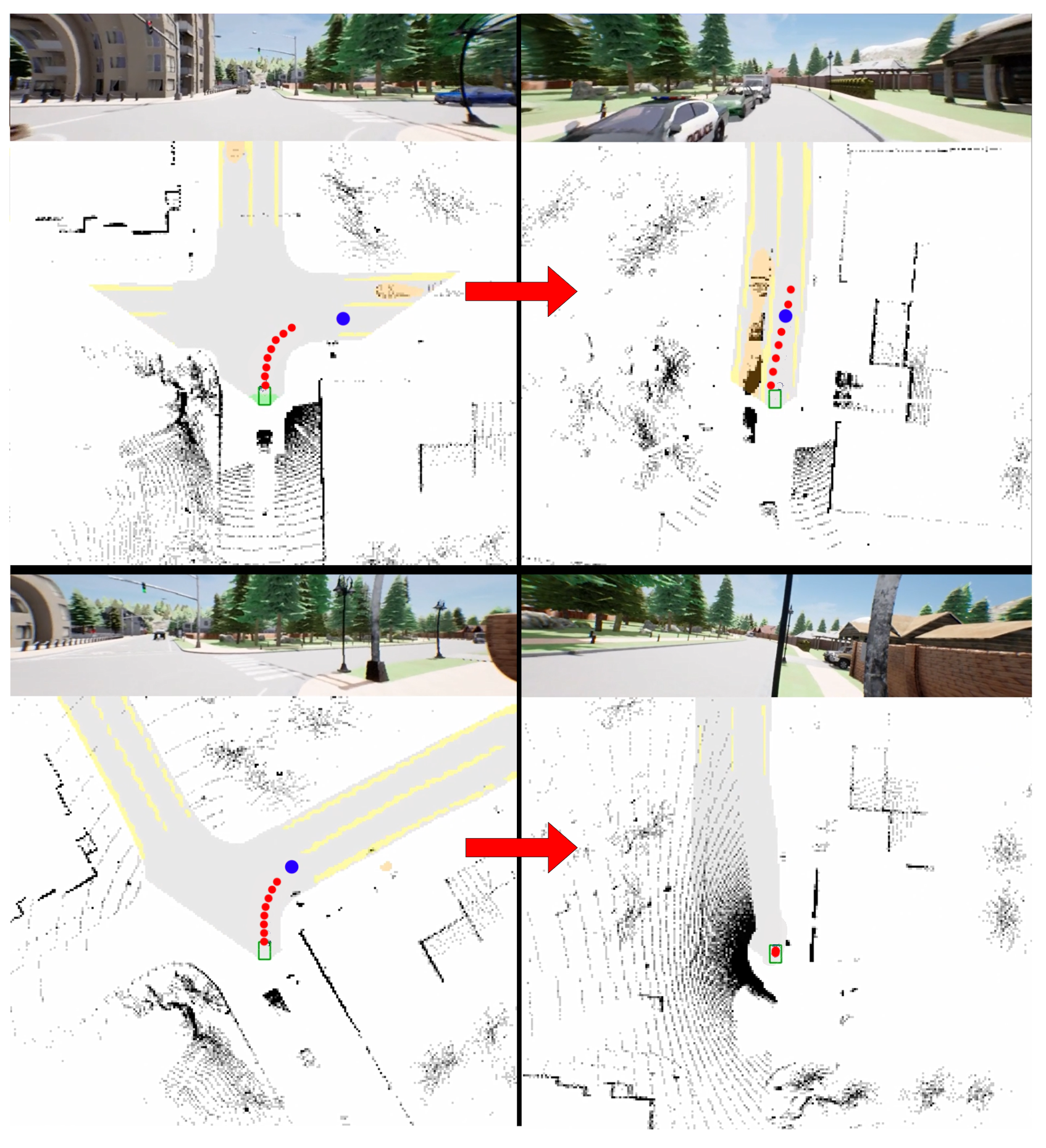}
\caption{A \texttt{Mini Cooper} making a right turn. \textbf{Top (Ours)}: MVAdapt executes a smooth, tight turn appropriate for the vehicle. \textbf{Bottom (Baseline)}: The baseline model over-rotates and hits the curb, failing the maneuver. The red dots represent the model’s output trajectory, and the blue dot indicates the target point.}
\label{fig:minicooper}
\end{figure}

\subsubsection{Performance Improvement through Few-Shot Adaptation}
For the exceptionally challenging case of the \texttt{Carla Cola Truck}, we observe a different regime from the average zero-shot setting above. Before fine-tuning, MVAdapt achieves a DS of 30.37 with RC of 100\% and IS of 0.30, indicating that the model can still finish routes but accumulates many infractions on this severe physical outlier. After fine-tuning on a small truck-specific dataset ($\sim$37K frames, about 3\% of the full training data), the DS rises to 61.9 and the IS to 0.62. This result supports a more precise claim: MVAdapt offers strong zero-shot transfer for many unseen vehicles, and it remains amenable to rapid few-shot calibration when the embodiment shift is exceptionally large.

This experiment highlights a practical deployment scenario. A general multi-vehicle model can be deployed first, and only vehicles that are far from the training distribution may require a short calibration phase instead of full re-training from scratch.

\subsection{Ablation Study}
To validate which components of MVAdapt are responsible for the gain, we include the ablation study from the thesis version of the work. Table~\ref{tab:ablation-results} compares the full model against three reduced variants on the Town05 Long benchmark: the TransFuser++ baseline without vehicle conditioning, a variant that adds physics information through simple concatenation only, and a variant that keeps the attention block but removes the learned physics encoder.

\begin{table}[t]
\centering
\caption{Ablation study on Town05 Long.}
\label{tab:ablation-results}
\resizebox{\columnwidth}{!}{%
\begin{tabular}{lccccc}
\toprule
\textbf{Model} & \textbf{Physics} & \textbf{Cross-Attn.} & \textbf{DS $\uparrow$} & \textbf{RC $\uparrow$} & \textbf{IS $\uparrow$} \\
\midrule
TransFuser++ baseline & No & No & 25.4 & 55.2 & 0.46 \\
Variant A (concat only) & Yes & No & 32.1 & 68.5 & 0.47 \\
Variant B (no physics encoder) & No & Yes & 28.9 & 60.1 & 0.48 \\
\textbf{MVAdapt} & Yes & Yes & \textbf{82.47} & \textbf{98.49} & \textbf{0.84} \\
\bottomrule
\end{tabular}}
\end{table}

The ablation shows that explicit vehicle information already helps compared with pure naive transfer, but the largest gain appears only when the learned physics encoder and the cross-attention fusion are used together. This supports the design choice that motivated MVAdapt: raw or weakly fused physical metadata is not sufficient, and the model must learn how vehicle embodiment changes which scene cues matter for waypoint prediction.

\subsection{Qualitative Analysis}
Qualitative observations from simulation runs further illustrate how MVAdapt adapts driving behavior to target vehicles. In various test routes, baseline agents often encountered problems that MVAdapt handled correctly:

\subsubsection{Maneuver Feasibility (Turning Radius)}
In one scenario with a large vehicle (\texttt{Tesla Cybertruck}), the URMA-based agent misjudged the turning radius and was blocked by another car, resulting in a scenario termination (Fig.~\ref{fig:cybertruck}). In contrast, MVAdapt, aware of the Cybertruck’s external dimension via its physics embedding, maintained a safer trajectory so that it could pass through in the identical situation. This led to MVAdapt completing the route successfully (RC 100\%), whereas URMA was prematurely terminated (RC \(\sim\)66\%). 

In another case with a very tiny car (\texttt{Mini Cooper}), the baseline BodyTransformer agent attempted a right turn at an intersection at a trajectory curvature appropriate for a midsize car: however, the Mini Cooper’s tighter turning ability meant it over-rotated and ended up partially off-road, triggering an infraction and getting stuck on the curb (Fig.~\ref{fig:minicooper}). MVAdapt considered the vehicle’s small size and agile steering, allowing it to execute a smoother, tighter turn without leaving the lane. The outcome was MVAdapt completing the turn and route (RC 100\%) with no infractions, whereas the baseline failed the route (RC 15\%). This result suggests that MVAdapt’s knowledge of the minimum turning radius constraint and vehicles’ turning reactions: it has learned to adjust planned paths to ensure they are physically achievable by the specific vehicle.

\subsubsection{Out-of-Bounds Predictions}
\begin{figure}
\centering
\includegraphics[width=\columnwidth]{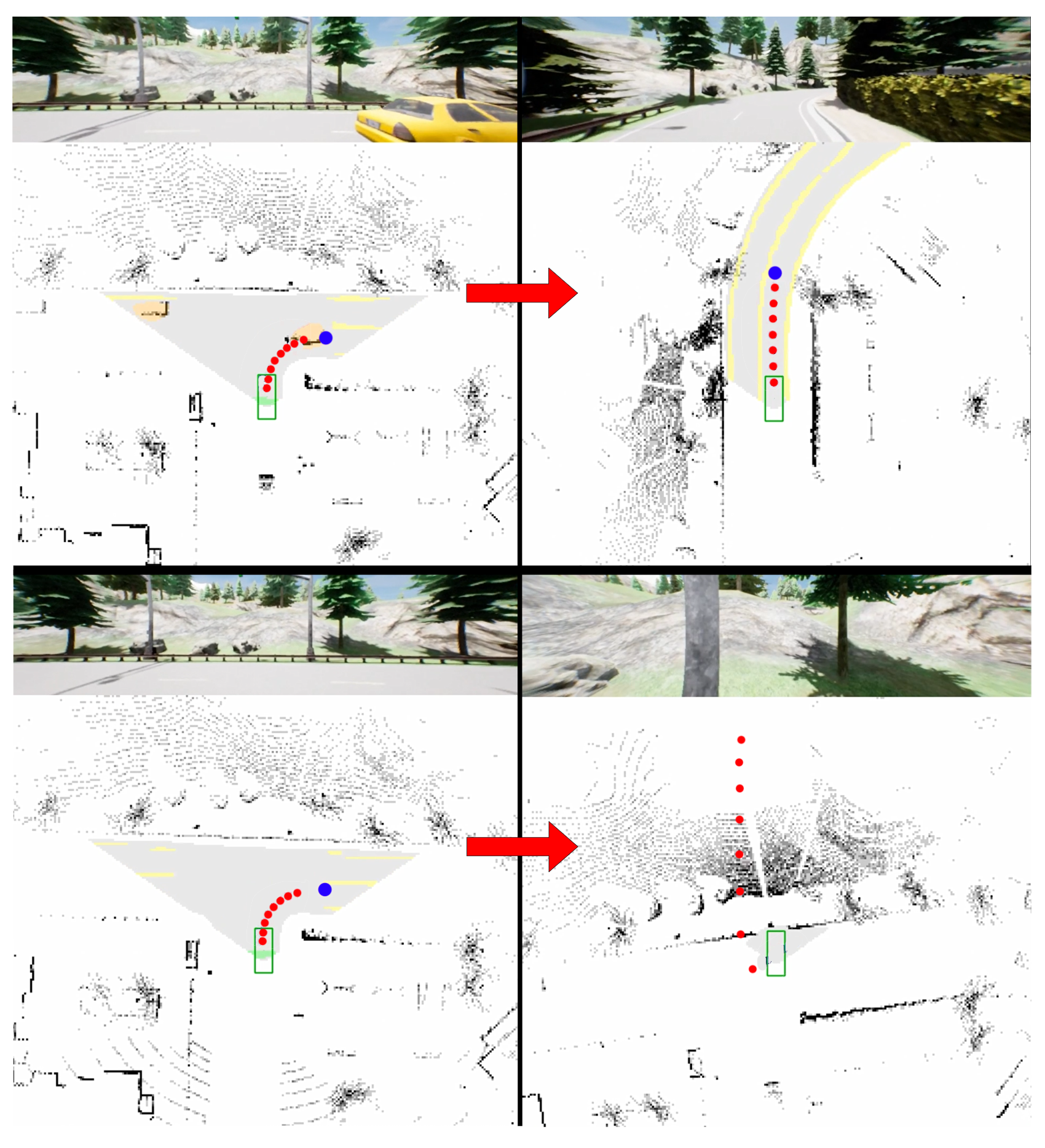}
\caption{Catastrophic failure during a right turn. \textbf{Top (Ours)}: MVAdapt generates a stable trajectory and safely completes the turn. \textbf{Bottom (Baseline)}: The baseline predicts an out-of-bounds trajectory, leading to an immediate failure. 
  The red dots represent the model’s output trajectory, and the blue dot indicates the target point.}
\label{fig:out-of-bound}
\end{figure}

We also noticed that the baseline models sometimes generate out-of-bounds predictions in certain situations. Fig.~\ref{fig:out-of-bound} presents the baseline BodyTransformer model suffering from a catastrophic failure during a right-turn maneuver. The BoT policy generated a trajectory that veered out of bounds mid-turn, leading to an immediate scenario termination. In contrast, MVAdapt executed the same tight right turn safely within lane boundaries. This case highlights MVAdapt’s ability to handle complex maneuvers for vehicles with different physical characteristics. The proposed method demonstrates the most accurate mapping between the physical properties of the target vehicle and its corresponding driving policy, resulting in stable trajectories.

\section{Conclusion} \label{sec:conclusion}
We presented \emph{MVAdapt}, a physics-conditioned adaptation framework for end-to-end autonomous driving across multiple vehicle types. By combining a frozen scene encoder with an explicit vehicle-physics branch and cross-attention fusion, MVAdapt improves cross-vehicle transfer in the CARLA Leaderboard benchmark while preserving strong in-distribution performance. The results support two main takeaways: first, the vehicle-domain gap is large enough to deserve explicit treatment; second, conditioning the policy on vehicle physics materially improves transfer to unseen vehicles. At the same time, our experiments also show that severe physical outliers remain challenging in zero-shot mode and benefit from a short few-shot calibration stage.

For future work, we plan to extend MVAdapt in several directions. First, we will explore adaptation to not just vehicle parameters but also \emph{hardware and actuator differences} (such as steering dynamics and latency), making the policy robust to a broader range of embodied differences. Second, testing on real-world driving data or actual vehicles will be crucial to validate whether the simulation findings carry over beyond CARLA. Third, implementing online adaptation is a promising future research topic. Similar to how a person adapts to a new vehicle, online adaptation during driving could become an effective mechanism for continuous multi-vehicle adaptation.

% %%%%%%%%%%%%%%%%%%%%%%%%%%%%%%%%%%%%%%%%%%%%%%%%%%%%%%%%%%%%%%%%%%%%%%%%%%%%%%%%
% \section*{APPENDIX}

% \section*{ACKNOWLEDGMENT}
%%%%%%%%%%%%%%%%%%%%%%%%%%%%%%%%%%%%%%%%%%%%%%%%%%%%%%%%%%%%%%%%%%%%%%%%%%%%%%%%

% References are important to the reader; therefore, each citation must be complete and correct. If at all possible, references should be commonly available publications.

\bibliographystyle{IEEEtran}
\bibliography{references}

\end{document}